\documentclass[10pt,twocolumn,letterpaper]{article}

\usepackage{cvpr}
\usepackage{times}
\usepackage{epsfig}
\usepackage{graphicx}
\usepackage{amsmath}
\usepackage{amssymb}
\DeclareMathOperator*{\argmin}{arg\,min}
\DeclareMathOperator{\Tr}{Tr}
\usepackage[]{mcode}
\usepackage{algorithmic}
\usepackage{algorithm}
\usepackage{booktabs}

\usepackage[pagebackref=true,breaklinks=true,letterpaper=true,colorlinks,bookmarks=false]{hyperref}
\cvprfinalcopy 

\def\cvprPaperID{1173} 

\ifcvprfinal\fi
\begin{document}

\title{Semantic Autoencoder for Zero-Shot Learning}

\author{Elyor Kodirov~~~~~Tao Xiang~~~~~Shaogang Gong\\
Queen Mary University of London, UK\\
{\tt\small \{e.kodirov, t.xiang, s.gong\}@qmul.ac.uk}
}

\maketitle

\begin{abstract}
Existing zero-shot learning (ZSL) models typically learn a projection function from a feature space to a semantic embedding space (e.g.~attribute space). However, such a projection function is only concerned with predicting the training seen class  semantic representation (e.g.~attribute prediction) or classification. When applied to test data, which in the context of ZSL contains different (unseen) classes without training data, a ZSL model typically suffers from the project domain shift problem. In this work, we present a novel solution to ZSL based on learning a Semantic AutoEncoder (SAE). Taking the encoder-decoder paradigm, an encoder aims to project a visual feature vector into the semantic space as in the existing ZSL models. However, the decoder exerts an additional constraint, that is, the projection/code must be able to reconstruct the original visual feature. We show that with this additional reconstruction constraint, the learned  projection function from the seen classes is able to generalise better to the new unseen classes. Importantly, the encoder and decoder are linear and symmetric which enable us to develop an extremely efficient learning algorithm. Extensive experiments on six benchmark datasets demonstrate that the proposed SAE outperforms significantly the existing ZSL models with the additional benefit of lower computational cost. Furthermore, when the SAE is applied to supervised clustering problem, it also beats the state-of-the-art. 
	 
\end{abstract}

\section{Introduction}
\begin{figure}[t!]
\centering
{%
\includegraphics[width=7.4cm, height=5.5cm]{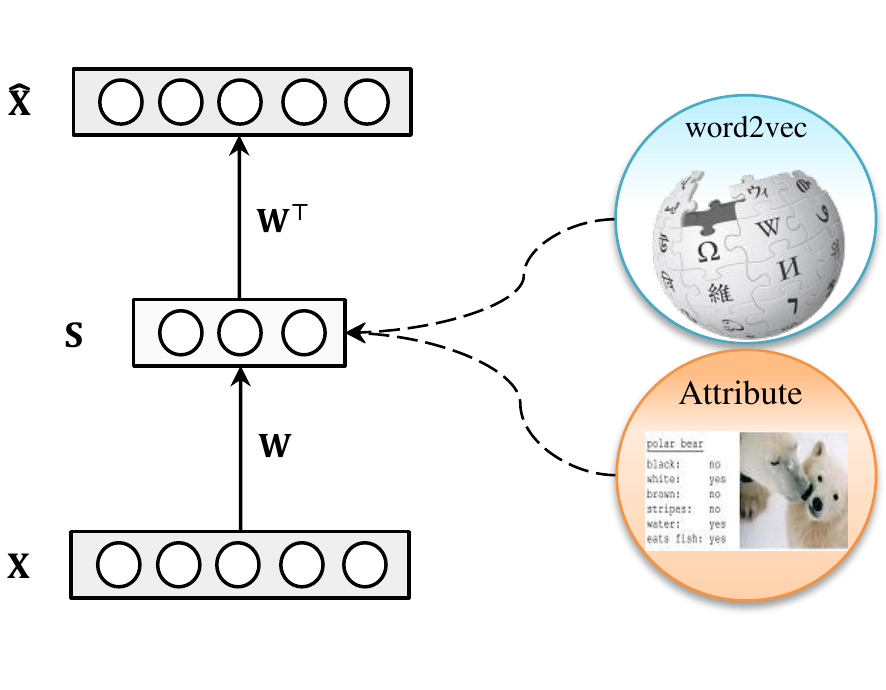}
\label{fig:concept}}
%
\caption{The proposed semantic autoencoder leverages the semantic side information such as  attributes and  word vector, while learning an encoder and a decoder. }
\end{figure}

A recent endeavour of computer vision research is to scale the visual recognition problem to large-scale. This is made possible by the emergence of  large-scale datasets such as ImageNet \cite{ILSVRC15} and the advances in deep learning techniques \cite{krizhevsky2012imagenet, sermanet2013overfeat, szegedy2015going, simonyan2014very}. However, scalability remains an issue because beyond daily objects, collecting image samples for rare and fine-grained object categories is difficult even with modern image search engines. 
Taking the ImageNet dataset for example, the popular large-scale visual recognition challenge (ILSVRC)
\cite{ILSVRC15} mainly focuses on the task of recognising
1K classes, a rather small subset of the full ImageNet dataset consisting of 21,814 classes with 14M images. This is because
many of the 21K object classes are only
composed of a handful of images including 296 classes with only one
image. 

Humans can identify approximately 30,000 basic object
categories~\cite{biederman1987recognition} and many more sub-classes,
e.g. breeds of dogs and combination of attributes and
objects. Importantly, humans are very good at recognising objects 
without seeing any visual samples. In machine learning, this is
considered as the problem of {\em zero-shot learning}
(ZSL). For example, a child would have no problem recognising a ``zebra''
if he/she has seen horses before and also learned that a ``zebra'' is
like a horse with black-and-white stripes. Inspired by humans'
ZSL ability, there is a recent surge of interest in machine learning
based ZSL for scaling up visual recognition to unseen object
classes without the need for additional data collection
\cite{zhang2015zero,rohrbach2011evaluating,frome2013devise,Romera-Paredes_2015,akata2015evaluation,Changpinyo_2016_CVPR, 
mensink2012metric,akata2013label,shigeto2015ridge,
Fu_2016_CVPR,fu2015zero,Norouzi_2014,bucher2016improving}.


Zero-shot recognition relies on the
existence of a labelled training set of seen classes and the 
knowledge about how each unseen class is semantically related to the seen
classes. Seen and unseen classes are usually related in a high dimensional
vector space, which is called semantic embedding space. Such a space can be a  semantic
attribute space \cite{lampert2014attribute} or a
semantic word vector space
\cite{frome2013devise,socher2013zero}. In the semantic
embedding space, the names of both seen and unseen classes are
embedded as vectors called  class prototypes
\cite{fu2014transductive}. The semantic relationships between classes
can then be measured by a distance, e.g. the prototypes of zebra and
horse should be close to each other. Importantly, the same space can
be used to project a feature representation of an object image, making
visual recognition possible. 
Specifically, most existing ZSL methods learn a projection (mapping)
function from a visual feature space to a semantic embedding space
using the labelled training visual data 
consisting of seen classes only. At test time for recognising
unseen objects, this mapping function is then used to
project the visual representation of an unseen class image
into the same semantic space where both seen and unseen classes reside. The task
of unseen class recognition is then realised by a simple nearest
neighbour (NN) search -- the class label of the test image is assigned
to the nearest unseen class prototype in the projected semantic space.


The training seen classes and testing unseen classes are
different. Although they can be considered as two overlapping domains
with some degrees of shared semantics, there exists significant domain
differences, e.g. the visual appearance of the same attributes can be
fairly different in unseen classes. Existing ZSL models mostly suffer
from the projection domain shift problem 
\cite{fu2015transductive}. This is, if the projection for visual
feature embedding is learned {\em only} from the seen classes, the
projections of unseen class images are likely to be misplaced
(shifted) due to
the bias of the training seen classes. Sometimes this shift could be
far away from the correct corresponding unseen class prototypes,
making the subsequent NN search inaccurate.


In this work, we present a novel approach to zero-shot learning based
on the encoder-decoder paradigm \cite{Ranzato07}. Specifically, an
encoder projects a visual feature representation of an image into a
semantic representation space such as an attributes space, similar to
a conventional ZSL model. However, we also consider the visual feature
projection as an input to a decoder which aims to reconstruct the
original visual feature representation. This additional reconstruction
task imposes a new constraint in learning the visual $\rightarrow$
semantic projection function so that the projection must also preserve
all the information contained in the original visual features,
i.e. they can be recovered by the decoder \cite{boureau2008sparse}. We
show that this additional constraint is very effective in mitigating
the domain shift problem. This is because although the visual
appearance of attributes may change from seen classes to unseen
classes, the demand for more truthful reconstruction of the visual
features is generalisable across seen and unseen domains, resulting in
the learned project function less susceptible to domain shift.

More precisely, we formulate a semantic autoencoder with the simplest
possible encoder and decoder model architecture
(Fig.~\ref{fig:concept}): Both have one linear projection to or from a 
shared latent embedding/code layer, and the encoder and decoder are
symmetric so that they can be represented by the same set of
parameters. Such a design choice is motivated by computational
efficiency -- the true potential of a ZSL model is when applied to
large-scale visual recognition tasks where computational speed is
essential.  Even with this simple formulation, 
solving the resultant optimisation problem efficiently is not
trivial. In this work, one such solver is developed whose complexity
is independent of the training data size therefore suitable for
large-scale problems.

Our semantic autoencoder differs from conventional autoencoder
\cite{Rumelhart:1986:LIR:104279.104293} in that the latent layer has
clear semantic meaning: It corresponds to the semantic space and is
subject to strong supervision. Therefore our model is not unsupervised. Beyond
ZSL learning, it can also be readily used for solving other 
problems where a discriminative low-dimensional representation is
required to cluster visually similar data points. To demonstrate its general applicability, our SAE model is formulated
for the supervised clustering problem
\cite{perrot2015regressive,Law_2016_CVPR}.


Our contributions are: (1) A novel semantic encoder-decoder model is
proposed for zero-shot learning. (2) We formulate a semantic
autoencoder which learns a low-dimensional semantic representation of
input data that can be used for data reconstruction. An efficient
learning algorithm is also introduced. (3) We show that the proposed
semantic autoencoder can be applied to other problems such as
supervised clustering. Extensive experiments are carried out on six
benchmarks for ZSL which show that the proposed SAE model achieves state-of-the-art performance on all the benchmarks.

\section{Related Work}
\label{sec:related work}

\noindent
\textbf{Semantic space}\quad A variety of zero-shot learning models have been proposed recently \cite{zhang2015zero,rohrbach2011evaluating,frome2013devise,Romera-Paredes_2015,akata2015evaluation,Changpinyo_2016_CVPR,
mensink2012metric,akata2013label,shigeto2015ridge,
Fu_2016_CVPR,fu2015zero,Norouzi_2014,bucher2016improving}. They use various semantic spaces. Attribute space is the most widely used. However, for large-scale problems, annotating attributes for each class becomes difficult. Recently, semantic word vector space has
started to gain popularity especially in large-scale zero-shot
learning~\cite{frome2013devise,socher2013zero}. Better
scalability is typically the motivation  as no manually
defined ontology is required and any
class name can be represented as a word vector for free. Beyond semantic attribute or word vector, direct 
learning from textual descriptions of categories has also been
attempted, e.g. Wikipedia articles
\cite{elhoseiny2013write,lei2015predicting}, sentence descriptions
\cite{reed2016learning}.

\noindent{\bf Visual $\rightarrow$ Semantic projection}\quad Existing ZSL models differ in how the visual space $\rightarrow$ semantic space projection function is established. They can be divided into three groups: (1) Methods in the first group learn a projection function from a visual feature space to a semantic space either using conventional regression or ranking models \cite{lampert2014attribute,akata2015evaluation} or via deep neural network regression or ranking \cite{socher2013zero,frome2013devise,reed2016learning,lei2015predicting}. (2) The second group chooses the reverse projection direction, i.e.~semantic $\rightarrow$ visual \cite{shigeto2015ridge,kodirov2015unsupervised}. The motivation is to alleviate the hubness problem that commonly suffered by nearest neighbour search in a high-dimensional space \cite{radovanovic2010hubs}. (3) The third group of methods learn an intermediate space where both the feature space and the semantic space are projected to  \cite{lu2015unsupervised,zhang2016zero,Changpinyo_2016_CVPR}. The encoder in our model is similar to the first group of models, whilst the decoder does the same job as the second group. The proposed semantic autoencoder can thus be considered as a combination of the two groups of ZSL models but with the added visual feature reconstruction constraint.

\noindent{\bf Projection domain shift}\quad The projection domain shift
problem in ZSL was first identified by Fu et 
al.~\cite{fu2015transductive}. In order to overcome this problem, a transductive multi-view
embedding framework was proposed together with label propagation on
graph which requires the access of all test data at once. Similar
transdutive approaches are proposed in
\cite{rohrbach2013transfer,kodirov2015unsupervised}.  This assumption
is often invalid in the context of ZSL because new classes typically
appear dynamically and unavailable before model learning. Instead of
assuming the access to all test unseen class
data for transductive learning, our model is based on inductive learning and it 
relies only enforcing the reconstruction constraint to the training data to counter domain shift. 


\noindent
\textbf{Autoencoder}\quad
There are many variants of autoencoders in the literature \cite{baldi1989neural,kavukcuoglu2009learning,lee2009unsupervised,vincent2008extracting,rifai2011contractive,rumelhart1988learning}. 
They can  be roughly divided into two groups which are (1)
undercomplete autoencoders and (2) overcomplete autoencoders. In
general, undercomplete  autoencoders are used to learn the underlying
structure of data and used for visualisation/clustering
\cite{xie2015unsupervised} like PCA. In contrast, overcomplete
autoencoders are used for classification based on the assumption that
higher dimensionnal features are better for classification
\cite{chenmarginalized,bengio2007greedy,bengio2009learning}. Our model
is an undercomplete autoencoder since a semantic space typically has
lower dimensionality than that of a visual feature space. All the
autoencoders above focus on learning features in a unsupervised
manner. On the contrary, our approach is supervised  while keeping the
main characteristic of the unsupervised autoencoders, i.e. the ability
to reconstruct the input signal.   

\noindent
\textbf{Semantic encoder-decoder} \quad An autoencoder is only one
realisation of the encoder-decoder paradigm. Recently deep
encoder-decoder has become popular for a variety of vision problems
ranging from image segmentation \cite{badrinarayanan2015segnet} to
image synthesis \cite{Attribute2Image,reed2016generative}. Among them,
a few recent works also exploited the idea of applying semantic
regularisation to the latent embedding space shared between the
encoder and decoder \cite{Attribute2Image,reed2016generative}. Our
semantic autoencoder can be easily extended for end-to-end deep
learning by formulating the encoder as a convolutional neural network
and the decoder as a deconvolutional neural network with a
reconstruction loss.


\noindent
\textbf{Supervised clustering}~~Supervised clustering methods exploit labelled clustering training dataset to learn a projection matrix that is shared by a test dataset unlike conventional clustering such as \cite{von2007tutorial,kodirov2015clustering}. There are different approaches of learning the  projection matrix: 1) metric learning-based methods that use similarity and dissimilarity constraints \cite{lajugie2014large,kostinger2012large,xing2003distance,
davis2007information}, and 2) regression-based methods that use `labels' \cite{perrot2015regressive,Law_2016_CVPR}. Our method is more closely related to the  regression-based methods, because  the training class labels are used to constrain the latent embedding space in our semantic autoencoder. We demonstrate in Sec~\ref{sec:exp_clustering} that, similar to the ZSL problem, by adding the reconstruction constraint, significant improvements can be achieved by our model on supervised clustering.

\section{Semantic Autoencoder}

\subsection{ Linear autoencoder}

We first introduce the formulation of a linear autoencoder and then proceed to extend it into a semantic one. In its simplest form, an autoencoder is linear and only has one hidden layer shared by the encoder and decoder. The encoder projects the input data into the hidden layer with a lower dimension and the decoder projects it back to the original feature space and aims to faithfully reconstruct the input data. Formally, given an input data matrix $\mathbf{X} \in \mathbb{R}^{d \times N}$ composed of $N$ feature vectors of $d$ dimensions as its columns, it is projected into a $k$-dimensional latent space with a projection matrix $\mathbf{W} \in \mathbb{R}^{k \times d}$, resulting in a latent representation $\mathbf{S} \in \mathbb{R}^{k \times N}$.  The obtained latent representation  is then projected back to the feature space with a projection matrix $\mathbf{W}^{*} \in \mathbb{R}^{d \times k}$ and becomes $\hat{\mathbf{X}} \in \mathbb{R}^{d \times N}$. We have $k < d$, i.e.~the latent representation/code reduces the dimensionality of the original data input. We wish that the reconstruction error is minimised, i.e.~$\hat{\mathbf{X}}$ is as similar as possible to $\mathbf{X}$. This is achieved by optimising against the following objective:
\begin{align}
\label{eq:objective_ae}
	 \min_{\mathbf{W},~ \mathbf{W}^{*}} \|\mathbf{X}-\mathbf{W}^{*}\mathbf{W}\mathbf{X}\|_F^2 
\end{align}


\subsection{Model Formulation}
A conventional autoencoder is unsupervised and the learned latent space has no explicit semantic meaning. With the proposed Semantic AutoEncoder (SAE), we assume that each data point also has a semantic representation, e.g.,~class label or attributes. To make the latent space in the autoencoder semantically meaningful, we take the simplest approach, that is, we force the  latent space $\mathbf{S}$ to be the semantic representation space, e.g., each column of $\mathbf{S}$ is now an attribute vector given during training for the corresponding data point. In other words, the latent space is not latent any more during training. The learning objective thus becomes:
\begin{align}
\label{eq:objective_main0}
 \min_{\mathbf{W},~ \mathbf{W}^{*}} \|\mathbf{X}-\mathbf{W}^{*}\mathbf{W}\mathbf{X}\|_F^2 ~~~s.t.~~~ \mathbf{W}\mathbf{X} = \mathbf{S}
\end{align}

\begin{figure}[t]
{
\includegraphics[width=7.5cm]{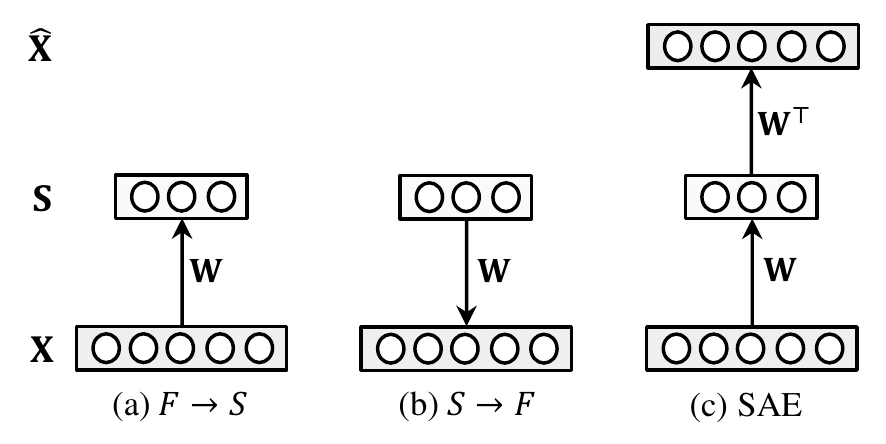}
\label{fig:disctSize}
}
\caption{Different ways of learning embedding space: (a) F $\rightarrow$ S, (b) S $\rightarrow$ F, and (c) Both (our SAE). 'F' -- Feature space, and 'S' -- Semantic space. }
\label{different_embedding}
\end{figure}

To further simplify the model, we consider tied weights \cite{boureau2008sparse}, that is: 
\begin{align}
	\mathbf{W}^{*} = \mathbf{W}^\top \nonumber
\end{align}
The learning objective is then rewritten as follows:
\begin{align}
\label{eq:objective_main}
\min_{\mathbf{W}} \|\mathbf{X}-\mathbf{W}^\top\mathbf{WX}\|_F^2 ~~~s.t.~~~ \mathbf{WX} = \mathbf{S}
\end{align}
Now we have only one projection matrix to estimate, instead of two (see Fig.~\ref{different_embedding}(c)).

\subsection{Optimisation}
To optimise the objective in Eq.~(\ref{eq:objective_main}), first we change Eq.~(\ref{eq:objective_main}) to the following form:
\begin{align}
\label{eq:objective_2}
\min_{\mathbf{W}} \|\mathbf{X} - \mathbf{W}^\top \mathbf{S}\|_F^2 ~~~s.t.~~~ \mathbf{WX} = \mathbf{S}
\end{align}
by substituting $\mathbf{WX}$ with $\mathbf{S}$. Solving an objective with a hard constraint such as $ \mathbf{WX} = \mathbf{S}$ is  difficult. Therefore, we consider to relax the constraint into a soft one and rewrite the objective as:
\begin{align}
\label{eq:objective_2_relax}
\min_{\mathbf{W}} \|\mathbf{X}-\mathbf{W}^\top \mathbf{S}\|_F^2 + \lambda \|\mathbf{WX} - \mathbf{S}\|_F^2
\end{align}
where $\lambda$ is a weighting coefficient that controls the importance of first and second terms, which correspond to the losses of the decoder and encoder respectively.
Now Eq.~(\ref{eq:objective_2_relax}) has a standard quadratic formulation, and it is convex function which has global optimal solution. 

To optimise it, we simply take a derivative of Eq.~(\ref{eq:objective_2_relax}) and set it zero. First, we re-organise  Eq.~(\ref{eq:objective_2_relax}) using trace properties  $\Tr(\mathbf{X}) = \Tr(\mathbf{X}^\top)$ and $\Tr(\mathbf{W}^\top\mathbf{S}) = \Tr(\mathbf{S}^\top\mathbf{W})$:
\begin{align}
\label{eq:objective_2_relax1}
\min_{\mathbf{W}} \|\mathbf{X}^\top - \mathbf{S}^\top \mathbf{W} \|_F^2 + \lambda \|\mathbf{WX} - \mathbf{S}\|_F^2
\end{align}

\noindent
Then, we can obtain the derivative of Eq. (\ref{eq:objective_2_relax1}) as follows:
\begin{align}
\label{eq:objective_2_relax2}
-\mathbf{S}(\mathbf{X}^\top-\mathbf{S}^\top \mathbf{W}) + \lambda (\mathbf{WX} - \mathbf{S}) \mathbf{X}^\top=0 \nonumber  \\
\mathbf{SS}^\top \mathbf{W} + \lambda \mathbf{WXX}^\top= \mathbf{SX}^\top + \lambda \mathbf{SX}^\top 
\end{align}

\noindent
If we denote $\mathbf{A} = \mathbf{SS}^\top$, $\mathbf{B} = \lambda \mathbf{XX}^\top$, and $\mathbf{C} = (1+ \lambda) \mathbf{SX}^\top$,  we have the following formulation:
\begin{align}
\label{eq:objective_2_relax3}
	\mathbf{AW} + \mathbf{WB} = \mathbf{C},
\end{align}
which is a well-known Sylvester equation which can be solved efficiently by the Bartels-Stewart algorithm \cite{bartels1972solution}. In MATLAB, it can be implemented with \textit{a single line} of code: \mcode{sylvester}\footnote{https://uk.mathworks.com/help/matlab/ref/sylvester.html}. Importantly, the complexity of Eq.~(\ref{eq:objective_2_relax3}) depends on the size of feature dimension ($\mathcal{O}(d^3)$), and not on the number of samples; it thus can scale to large-scale datasets. 
Algorithm 1 shows a 6-line MATLAB implementation of our solver.  

\begin{table}[t!]
\centering
\label{my-label}
\begin{tabular}{@{}l@{}}
\toprule
\textbf{Algorithm 1} SAE in MATLAB ~~~~~~~~~~~~~~~~~~~~~~~~~~~~~~~~~~~\\ \midrule
\begin{lstlisting}
  function W = SAE(X,S,lambda)    
    % SAE - Semantic AutoEncoder
    % Input:
    %    X: dxN data matrix.
    %    S: kxN semantic matrix.
    %    lambda: regularisation parameter.
    %
    % Return: 
    %    W: kxd projection matrix.

    A = S*S';
    B = lambda*X*X';
    C = (1+lambda)*S*X';
    W = sylvester(A,B,C);
  end
\end{lstlisting}                           \\ \bottomrule
\end{tabular}
\end{table}


\section{Generalisation}

\subsection{Zero-Shot Learning}
\label{sec:app_ZSL}
\noindent
\noindent
\textbf{Problem definition~~} Let $\mathbf{Y} = \{\mathbf{y}_1, ...~, \mathbf{y}_s\}$ and ${\mathbf{Z}} = \{\mathbf{z}_1, ...~, \mathbf{z}_u\}$ denote a set of $s$ seen and $u$ unseen class labels, and they are disjoint  $\mathbf{Y} \cap \mathbf{Z} = \varnothing$. Similarly ${\mathbf{S}_Y} = \{ \mathbf{s}_1, ...~, \mathbf{s}_s\}\in \mathbb{R}^{s \times k}$  and ${\mathbf{S}_Z} = \{ \mathbf{s}_1, ...~, \mathbf{s}_u\}\in \mathbb{R}^{u \times k}$ denote the corresponding seen and unseen class semantic representations (e.g.~$k$-dimensional attribute vector). Given training data with $N$ number of samples $\mathbf{X}_Y = \{(\mathbf{x}_i, \mathbf{y}_i, \mathbf{s}_i)\} \in \mathbb{R}^{d \times N}$, where $\mathbf{x}_i$ is a $d$-dimensional visual feature vector extracted from the $i$-th training image from one of the seen classes, zero-shot learning aims to learn a classifier $f: \mathbf{X}_Z\rightarrow \mathbf{Z}$ to predict the label of the image coming from unseen classes, where $\mathbf{X}_Z = \{(\mathbf{x}_i, \mathbf{z}_i, \mathbf{s}_i)\}$ is the test data and  $\mathbf{z}_i$ and $\mathbf{s}_i$ are unknown.

\noindent
\textbf{SAE for zero-shot learning~~}
\noindent
Given semantic representation $\mathbf{S}$ such as attributes, and the training data $\mathbf{X_Y}$, using our SAE, we first learn the encoder $\mathbf{W}$ and decoder $\mathbf{W}^\top$ by Algorithm 1. Subsequently, zero-shot classification can be performed in two spaces:

\begin{itemize}
\item[1)] With the encoder projection matrix $\mathbf{W}$: We can embed a new test sample $\mathbf{x}_i \in \mathbf{X}_Z$ to the semantic space by $\mathbf{\hat{s}}_i = \mathbf{W}\mathbf{x}_i$. After that, the classification of the test data in the semantic space can be achieved by simply calculating the distance between the estimated semantic representation $\mathbf{s}_i$ and the projected prototypes $\mathbf{S}_Z$:
\begin{equation}
	\Phi(\mathbf{x}_i) = \argmin_j D(\mathbf{\hat{s}}_i,\mathbf{S}_{Z_j})
\end{equation} 

\noindent
where $\mathbf{S}_{Z_j}$ is $j$-th prototype attribute vector of the $j$-th unseen class, $D$ is a distance function, and $\Phi(\cdot)$ returns the class label of the sample.

\item [2)] With the decoder projection matrix $\mathbf{W}^\top$: Similarly, we can embed the prototype representations to the visual feature space by $\hat{\mathbf{x}}_i=\mathbf{W}^T\mathbf{s}_i$  where $\mathbf{s}_i \in \mathbf{S}_Z$ and $\hat{\mathbf{x}}_i \in \hat{\mathbf{X}}_Z$ is the projected prototype. Then, the classification of the test data in the feature space can be achieved by calculating the distance between the  feature representation $\mathbf{x}_i$ and the prototype projections in the feature space $\mathbf{\hat{X}}_Z$:
\begin{equation}
	\Phi(\mathbf{x}_i) = \argmin_j D(\mathbf{{x}}_i,\mathbf{\hat{X}}_{Z_j})
\end{equation} 
where  $\mathbf{\hat{X}}_{Z_j}$ is $j$-th unseen class prototype projected in the feature space.

\end{itemize}

\noindent
In our experiments we found that the two testing strategies yield very similar results (see Sec.~\ref{sec:exp_zsl}). We report results with both strategies unless otherwise specified.




\subsection{Supervised Clustering}

For supervised clustering we are given a set of training data with class labels only, and a test set that share the same feature representation as the training data and need to be grouped into clusters. Let $\mathbf{Y} = \{\mathbf{y}_1, ...~, \mathbf{y}_s\}$ be a set of $s$ training class labels. Denote ${\mathbf{S}_Y} = \{ \mathbf{s}_1, ...~, \mathbf{s}_s\}\in \mathbb{R}^{s \times k}$  as the corresponding  semantic representations. Given a training data set with $N$ number of samples $\mathbf{X}_Y = \{(\mathbf{x}_i, \mathbf{y}_i, \mathbf{s}_i)\} \in \mathbb{R}^{d \times N}$, where $\mathbf{x}_i$ is a $d$-dimensional visual feature vector extracted from the $i$-th training image we aim to learn a projection function $f: \mathbf{X}_Y \rightarrow \mathbf{S}_Y$ from the training data and then apply the same projection function  to a set of test data $\mathbf{X}_Z$ before clustering can be carried out. 

Using our SAE, the projection function is our encoder $\mathbf{W}$. With only the training class label, the semantic space is the label space, that is, $\mathbf{s}_i$ is an one-hot class label vector with only the element corresponding to the image class assuming the value $1$, and all other elements set to $0$. After the test data is projected into the training label space, we use $k$-means clustering as in existing work  \cite{perrot2015regressive,Law_2016_CVPR} for fair comparison. The demo code of our model is available at \url{https://elyorcv.github.io/projects/sae}.
 

\subsection{Relations to Existing Models}
\label{sec:analysis}
\noindent
\textbf{Relation to ZSL models~~}
Many existing ZSL models learn a projection function from a visual feature space to a semantic space (see Fig.~\ref{different_embedding}(a)). If the projection function is formulated as linear ridge regression as follows: 
\begin{align}
\label{eq:objective_regression}
\min_{\mathbf{W}} \|\mathbf{WX} - \mathbf{S}\|_F^2 + \lambda \|\mathbf{W}\|_F^2,
\end{align}
we can see that comparing  Eq.~(\ref{eq:objective_regression}) with  Eq.~(\ref{eq:objective_2_relax}), this is our encoder with an additional regularisation term on the project matrix $\mathbf{W}$. 

Recently, \cite{shigeto2015ridge} proposed to reverse the projection direction: They project the semantic prototypes into the features space:
\begin{align}
\label{eq:reverse_regression}
\min_{\mathbf{W}} \|\mathbf{X} - \mathbf{W^\top S}\|_F^2 + \lambda \|\mathbf{W}\|_F^2
\end{align}
\noindent
so this is the decoder of our SAE but again with the regularisation term to avoid overfitting (see Fig.~\ref{different_embedding}(b)). 

Our approach can thus be viewed as the combination of both models when ridge regression is chosen as the project function and without considering the $\|\mathbf{W}\|_F^2$ regularisation. This regularisation is unnecessary in our model due to the symmetric encoder-decoder design -- since $\mathbf{W}^{*} = \mathbf{W}^\top$, the norm of the encoder projection matrix $\|\mathbf{W}\|_F^2$ cannot be big because it will then produce large-valued projections in the semantic space, and after being multiplied with the large-norm decoder project matrix, will result in bad reconstruction. In other words, the regularisation on the norm of the projection matrices have been automatically taken care of by the reconstruction constraint \cite{boureau2008sparse}.

\noindent
\textbf{Relation to supervised clustering models~~}
\noindent
Recently, \cite{perrot2015regressive,Law_2016_CVPR} show that regression can be used to learn a mahalanobis distance metric for supervised clustering. Specifically, given data $\mathbf{X}$ with corresponding labels, the so-called `encoded labels' $\mathbf{S}$ are generated and normalised as $\mathbf{S} = \mathbf{S}(\mathbf{S}'\mathbf{S})^{-1/2} \in \mathbb{R}^{s\times N}$, where $s$ is the number of training labels \cite{Law_2016_CVPR}. Then linear regression  is employed to obtain a projection matrix $\mathbf{W}$ for projecting the data from the feature space to the label space. At test time, $\mathbf{W}$ is applied to test data. Then, $k$-means clustering is applied to the projected data. Again, these models can be considered as the encoder of our SAE. We shall show that with the decoder and the additional reconstruction constraint, the learned code and distance metric become more meaningful, yielding superior clustering performance on the test data. 

\section{Experiments}
\label{sec:exp}

\subsection{Zero-Shot Learning}
\label{sec:exp_zsl}

\begin{table}[h!]
\centering
\small
\begin{tabular}{|l|c|l|c|c|}
\hline
Dataset       & \#instances         	 &	 SS & SS-D & \# seen/unseen  \\ \hline \hline
AwA \cite{lampert2014attribute}          & 30,475                & A  & 85              & ~~~40 / 10~~~                \\
CUB \cite{wah2011caltech} 				 & 11,788                & A  & 312             & ~150 / 50~~~               \\
aP\&Y \cite{farhadi2009describing}       & 15,339                & A  & 64              & ~~20 / 12~~~                \\
SUN \cite{patterson2014sun} 		  	 & 14,340                & A  & 102             & ~645 / 72 $^{(*)}$               \\ \hline \hline
ImNet-1 \cite{ILSVRC15} 				 &          1,2 $\times10^6$              & W  & 1,000            & ~800 / 200              \\
ImNet-2 \cite{ILSVRC15} & 218,000 & W  & 1,000            & 1,000 / 360             \\ 
\hline
\end{tabular}
\caption{Benchmark datasets for evaluation. Notation: `SS' -- semantic
  space, `SS-D' -- the dimension of semantic space, `A' -- attribute,
  and `W' -- word vector. $^{(*)}$ -- another split of 707/10 is also
  used for SUN \cite{jayaraman2014zero,zhang2016zero}.} 
\label{dataset}
\end{table}

\begin{table*}[ht!]
\small
\centering
\begin{tabular}{|l|c|c|c|c|c||l|c|c|c|}
\hline
\multicolumn{6}{|c||}{Small-scale datasets}                                                     & \multicolumn{4}{c|}{Large-scale datasets}                             \\ \hline
\multicolumn{1}{|c|}{Method}              & SS    					   & AwA           & CUB           & aP\&Y & SUN             & \multicolumn{1}{c|}{Method} & SS    & ImNet-1    & ImNet-2    \\ \hline \hline
DAP                        \cite{lampert2014attribute}      & A   & 60.1          & -             & 38.2  & 72.0~$|$~44.5          & Rohrbach \etal                  	    \cite{rohrbach2011evaluating}& W     & 34.8          & --             \\
ESZSL                      \cite{Romera-Paredes_2015}       & A   & 75.3          & 48.7          & 24.3  & 82.1~$|$~18.7          & Mensink \etal                          \cite{mensink2012metric}	 & W	 & 35.7   & --             \\
SSE                 	   \cite{zhang2015zero} 		    & A   & 76.3          & 30.4          & 46.2  & 82.5~$|$~~--~~~~       & DeViSE                                 \cite{frome2013devise} 	 & W	 & 31.8   & 12.8          \\
SJE                        \cite{akata2015evaluation}       & A+W & 73.9          & 50.1          & -     & ~~--~~~~$|$~56.1       & ConSE                      		    \cite{Norouzi_2014}  		 & W	 & 28.5          & 15.5             \\
JLSE                       \cite{zhang2016zero}             & A   & 80.5          & 41.8          & 50.4  & 83.8~$|$~~--~~~~       & AMP                      			    \cite{fu2015zero}  		 & W	 & 41.0          & 13.1          \\
SynC$^{struct}$            \cite{Changpinyo_2016_CVPR}      & A & 72.9            & 54.4          & -     & ~~--~~~~$|$~62.7       & SS-Voc                 			     \cite{Fu_2016_CVPR}  		 & W	 & --             & 16.8          \\
MLZSC                      \cite{bucher2016improving}       & A   & 77.3          & 43.3          & 53.2  & 84.4~$|$~~--~~~~       & PST                		            \cite{rohrbach2013transfer}  & W	 & 34.0          & --          \\  
DS-SJE                     \cite{reed2016learning}          & A/D & -             & 50.4/56.8     & -     & --~~~~$|$~~--~~        
 &                  		                                         & 	 &               &            \\ AMP                        \cite{bucher2016improving}       & A+W & 66.0          & -             & -     & --~~~~$|$~~--~~        &  &  		 &           &           \\  
DeViSE                     \cite{frome2013devise}           & A/W & 56.7/50.4     & 33.5          & -     & --~~~~$|$~~--~~        &                 		                & 		 &           &           \\  
RRZSL                      \cite{shigeto2015ridge}          & A   & 80.4          & 52.4          & 48.8  & 84.5~$|$~~--~~~~       & &   &              &           \\  
Ba \etal				   \cite{lei2015predicting}         & A/W & 69.3/58.7     & 34.0          & -     & --~~~~$|$~~--~~        & &   &              &           \\  
MTMDL 				       \cite{yang2015unified}           & A/W & 63.7/55.3     & 32.3          & -     & --~~~~$|$~~--~~        & &   &              &           \\  
SS-voc 			           \cite{Fu_2016_CVPR}              & A/W & 78.3/68.9     & -             & -     & --~~~~$|$~~--~~        & &   &              &           \\  
  			             \hline \hline
SAE ($\mathbf{W}$)                 	& A & \textbf{84.7} & \textbf{61.4} & \textbf{55.4}  & \textbf{91.0~$|$~65.2} & SAE ($\mathbf{W}$)                     & W & \textbf{46.1} & \textbf{26.3} \\ 
SAE ($\mathbf{W^\top}$)                	& A & \textbf{84.0} & \textbf{60.9} & \textbf{54.8}  & \textbf{91.5~$|$~65.2} & SAE ($\mathbf{W^\top}$)                      & W & \textbf{45.4} & \textbf{27.2} \\\hline
\end{tabular}
\caption{Comparative ZSL classification accuracy (\%, hit@5 for
  large-scale datasets). For SS (Semantic Space), `/' means `or' and
  `+' means `and'. For CUB, 10 sentence description per image are also
  used in \cite{reed2016learning} as input to a language model
  (word-CNN-RNN) to compute semantic space (`D'). For the SUN dataset,
  the results are for the 707/10 and 645/72 splits respectively,
  separated by `$|$'. `-' means that no reported results are
  available. $\mathbf{W}$ parametrises the projection function of the encoder and $\mathbf{W}^\top$ the decoder.}  
\label{main_zsl_result}
\end{table*}

\noindent
\textbf{Datasets}~~Six benchmark datasets are used. Four of them are small-scale datasets: Animals with Attributes (AwA) \cite{lampert2014attribute}, CUB-200-2011 Birds (CUB) \cite{wah2011caltech}, aPascal\&Yahoo (aP\&Y) \cite{farhadi2009describing}, and SUN Attribute (SUN) \cite{patterson2014sun}. The two large-scale ones are ILSVRC2010 \cite{deng2009imagenet} (ImNet-1), and ILSVRC2012/ILSVRC2010 \cite{ILSVRC15} (ImNet-2). In ImNet-2, as in \cite{Fu_2016_CVPR}, the 1,000 classes of ILSVRC2012 are used as seen classes, while 360 classes of ILSVRC2010, which are not included in ILSVRC2012, for unseen classes. The summary of these datasets is given in Table \ref{dataset}.

\noindent
\textbf{Semantic spaces}~~We use attributes as the semantic space for the small-scale datasets, all of which provide  the attribute annotations. Semantic word vector representation is used for large-scale datasets. We train a skip-gram text model  on a corpus of 4.6M Wikipedia documents to obtain the word2vec\footnote { https://code.google.com/p/word2vec/ } \cite{mikolov2013distributed,mikolov2013efficient} word vectors. 

\noindent
\textbf{Features}~~All recent ZSL methods use visual features extracted by deep convolutional neural networks (CNNs). In our experiments, we use GoogleNet features \cite{szegedy2015going} which is the 1024D activation of the final pooling layer as in \cite{akata2015evaluation}. The only exception is for ImNet-1: For fair comparison with published results,  we use Alexnet \cite{krizhevsky2012imagenet} architecture, and train it from scratch using the 800 seen classes, resulting in 4096D visual feature vectors computed using the FC7 layer. 

\noindent
\textbf{Parameter settings}~~Our SAE model has only one free parameter: $\lambda$ (see Eq.~(\ref{eq:objective_2_relax})). As in \cite{zhang2016zero}, its values is set by class-wise cross-validation using the training data. The dimension of the embedding (middle) layer always equals to that of the semantic space. Only SUN dataset has multiple splits. We use the same 10 splits used in  \cite{Changpinyo_2016_CVPR}, and report the average performance.

\noindent
\textbf{Evaluation metric}~~For the small-scale datasets, we use multi-way classification accuracy as in previous works, while for the large-scale datasets flat hit@K classification accuracy is used as in \cite{frome2013devise}. hit@K means that the test image is classified to a `correct label' if it is among the top K labels. We report hit@5 accuracy  as in other works, unless otherwise stated. 

\noindent
\textbf{Competitors~~}  14 existing ZSL models are selected for the small-scale datasets and 7 for the large-scales ones (much fewer existing works reported results on the large-scale datasets). The selection criteria are: (1) recent work: most of them are published in the past two years; (2) competitiveness: they clearly represent the state-of-the-art; and (3) representativeness: they cover a wide range of models (see Sec.~\ref{sec:related work}).

\noindent \textbf{Comparative evaluation}\quad
From the results in Table \ref{main_zsl_result} we can make the following observations: (1) Our SAE model achieves the best results on all 6 datasets. (2) On the small-scale datasets, the gap between our model's results to the strongest competitor ranges from 3.5\% to 6.5\%. This is despite the fact that most of the compared models use far complicated nonlinear models and some of them use more than one semantic space.  (3) On the large-scale datasets, the gaps are even bigger: On the largest ImNet-2, our model improves over the state-of-the-art SS-Voc \cite{Fu_2016_CVPR} by 8.8\%. (4) Both the encoder and decoder projection functions  in our SAE model (SAE ($\mathbf{W}$) and SAE ($\mathbf{W^\top}$) respectively) can be used for effective ZSL. The encoder projection function seems to be slightly better overall.

\begin{table}[h!]
\centering
\begin{tabular}{|c|c|c|c|c|c|c|}
\hline 
Projection                    & AwA             & CUB             & aP\&Y           & SUN              \\ \hline \hline
F $\rightarrow$ S 		      & 60.6            & 41.1            & 30.5            & 71.5             \\
F $\leftarrow$ S			  & 80.4            & 52.4            & 48.8            & 84.5             \\ \hline \hline
SAE                           & \textbf{84.7}   & \textbf{61.4}   & \textbf{55.4}   & \textbf{91.0}    \\ \hline
\end{tabular}
\caption{The importance of adding the reconstruction constraint. Both compared methods are based on ridge regression and differ in the projection direction between the visual  and semantic  spaces. Attributes are used. The encoder is used. }
\label{different_embedding_table}
\end{table}

\noindent \textbf{Ablation study}\quad
The key strength of our model comes from the additional reconstruction constraint in the autoencoder formulation. Since most existing ZSL models use more sophisticated projection functions than our linear mapping, in order to evaluate how important this additional constraint is, we consider ZSL baselines that use the same simple projection functions as our model.  As discussed in Sec.~\ref{sec:analysis}, without the constraint both the encoder and decoder can be considered as conventional ZSL models with linear ridge regression as projection function, and they differ only in the project directions. Table \ref{different_embedding_table} shows than, when the projection function is the same, adding the additional reconstruction constraint makes a huge difference. Note that comparing to the state-of-the-art results in Table \ref{main_zsl_result}, simple ridge regression is competitive but clearly inferior to the best models due to its simple linear projection function. However, when the two models are combined in our SAE, we obtain a much more powerful model that beats all existing models. 
  

\noindent \textbf{Generalised Zero-Shot Learning}\quad
\noindent
Another ZSL setting that emerges recently is the generalised setting under which the test set contains data samples from both the seen and unseen classes.   We follow the same setting of \cite{gzsl}. Specifically, we hold out 20\% of the data samples from the seen classes and mix them with the data samples from the unseen classes. 
The evaluation metric is now Area Under Seen-Unseen accuracy Curve (AUSUC), which measures how well a zero-shot learning method can trade-off between recognising data from seen classes and that of unseen classes \cite{gzsl}. The upper bound of this metric is 1. 
The results on AwA and CUB are presented in
Table \ref{gzsl_result}  comparing our model with 5 other alternatives. We can see that on AwA, our model is slightly worse than the state-of-the-art method SynC$^{struct}$ \cite{Changpinyo_2016_CVPR}. However, on the more challenging CUB dataset, our method significantly outperforms the competitors.

\begin{table}[]
\small
\centering
\begin{tabular}{|l|c|c|c|}
\hline
Method    & AwA   & CUB    \\ \hline \hline
DAP \cite{lampert2014attribute}       & 0.366 & 0.194           \\
IAP \cite{lampert2014attribute}       & 0.394 & 0.199           \\
ConSE \cite{Norouzi_2014}   		  & 0.428 & 0.212           \\
ESZSL \cite{Romera-Paredes_2015}      & 0.449 & 0.243			\\
SynC$^{struct}$ \cite{Changpinyo_2016_CVPR} & \textbf{0.583} & 0.356           \\ \hline \hline
SAE      & 0.579 & \textbf{0.448}           \\ \hline
\end{tabular}
\caption{Comparative evaluation on generalised zero-shot learning on AwA and CUB. Encoder is used. }
\label{gzsl_result}
\end{table}

\noindent
\textbf{Computational cost}~~We evaluate the computational cost of our method in comparison with three linear ZSL models SSE \cite{zhang2015zero}, ESZSL \cite{Romera-Paredes_2015} and AMP \cite{fu2015zero} which are among the more efficient existing ZSL models. Table \ref{my-label} shows that for model training, our SAE is at least 10 times faster. For testing, our model is still the fastest, although ESZSL is close. 
 
\begin{table}[h]
\small
\centering
\begin{tabular}{|l|c|c|} \hline
Method    					   	      & Training   & Test   \\ \hline \hline
SSE   \cite{zhang2015zero} 		  & 1312       		  & 9.20      \\
ESZSL \cite{Romera-Paredes_2015}  & 16            	  & 0.08      \\
AMP   \cite{fu2015zero}   		  & 844           	  & 0.23      \\ \hline \hline
SAE  							  & 1.3               & 0.07      \\ \hline
\end{tabular}
\caption{Evaluating the computational cost (in second) on AwA. Encoder is used.}
\label{my-label}
\end{table}

%

\subsection{Supervised Clustering}
\label{sec:exp_clustering}


\begin{table*}[ht]
\footnotesize
\centering
\begin{tabular}{|l|c|c|c|c|c|c|c|c|}
\hline
Method                     & SAE          & $L_2$          & Xiang \etal \cite{xing2003distance} & Lajugie \etal \cite{lajugie2014large} & KISSME \cite{kostinger2012large} & ITML \cite{davis2007information} & LMNN \cite{weinberger2009distance} & MLCA \cite{Law_2016_CVPR}         \\ \hline
Test loss                  & \textbf{0.01} & 3.0                & 0.7          & 0.11           & 0.07   & 0.08 & 3.0  & 0.07           \\
Training Time (in second) & {0.020}           & NT        & 4378         & 336            & 0.5    & 370  & 4    & \textbf{0.004} \\ \hline
\end{tabular}
\caption{Supervised clustering results on synthetic data with clusters of same size. 'NT'-- No Training,  $L_2$--Euclidean distance. Encoder is used.}
\label{sc_same_size}
\end{table*}

\begin{table*}[ht]
\footnotesize
\centering
\begin{tabular}{|l|c|c|c|c|c|c|c|c|}
\hline
Method                     & SAE          & $L_2$           & Xiang \etal \cite{xing2003distance} & Lajugie \etal \cite{lajugie2014large} & KISSME \cite{kostinger2012large} & ITML \cite{davis2007information} & LMNN \cite{weinberger2009distance} & MLCA \cite{Law_2016_CVPR}         \\ \hline
Test loss                  & \textbf{0.01}          & 3.0                & 3.0          & 0.09           & 2.02   & 3.0   & 3.0  & 0.09           \\
Training Time (in second) & {0.026}         & NT        & 21552        & 2462           & 2      & 1260  & 11   & \textbf{0.005} \\ \hline
\end{tabular}
\caption{Supervised clustering results on synthetic data  with clusters of different sizes and with noise (lower is better). Encoder is used.}
\label{sc_diff_size}
\end{table*}

\begin{table*}[t!]
\footnotesize
\centering
\begin{tabular}{|l|c|c|c|c|c|c|}
\hline
Method                     & SAE                              & Lajugie \etal \cite{lajugie2014large} & KISSME \cite{kostinger2012large} & ITML \cite{davis2007information} & LMNN \cite{weinberger2009distance} & MLCA \cite{Law_2016_CVPR}         \\ \hline
Test loss                  & \textbf{1.19$\pm$0.01}            & 1.38$\pm$0.02                         & 1.59$\pm$0.02                    & 1.50$\pm$0.02                    & 1.79$\pm$0.02                      & 1.29$\pm$0.01              \\
Training Time                   & 93 seconds                        & 5 days                                & 11 minutes                       & 2 hours                          & 1 day                              & \textbf{39 seconds}            \\ \hline
\end{tabular}
\caption{Supervised clustering (segmentation) results on Oxford Flowers. Encoder is used.}
\label{flower}
\end{table*}

\begin{figure*}[ht!]
\centering
{%
\includegraphics[width=4.0cm]{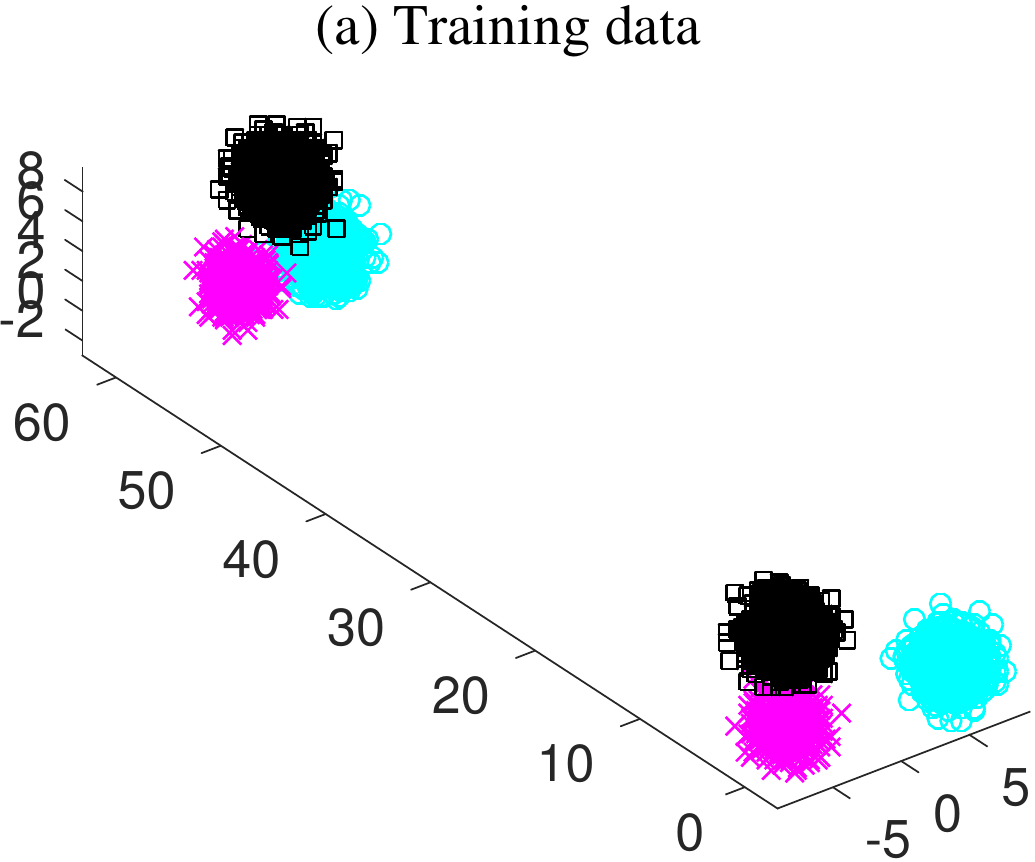} 
\label{fig:disctSize}
}%
{%
\includegraphics[width=4.0cm]{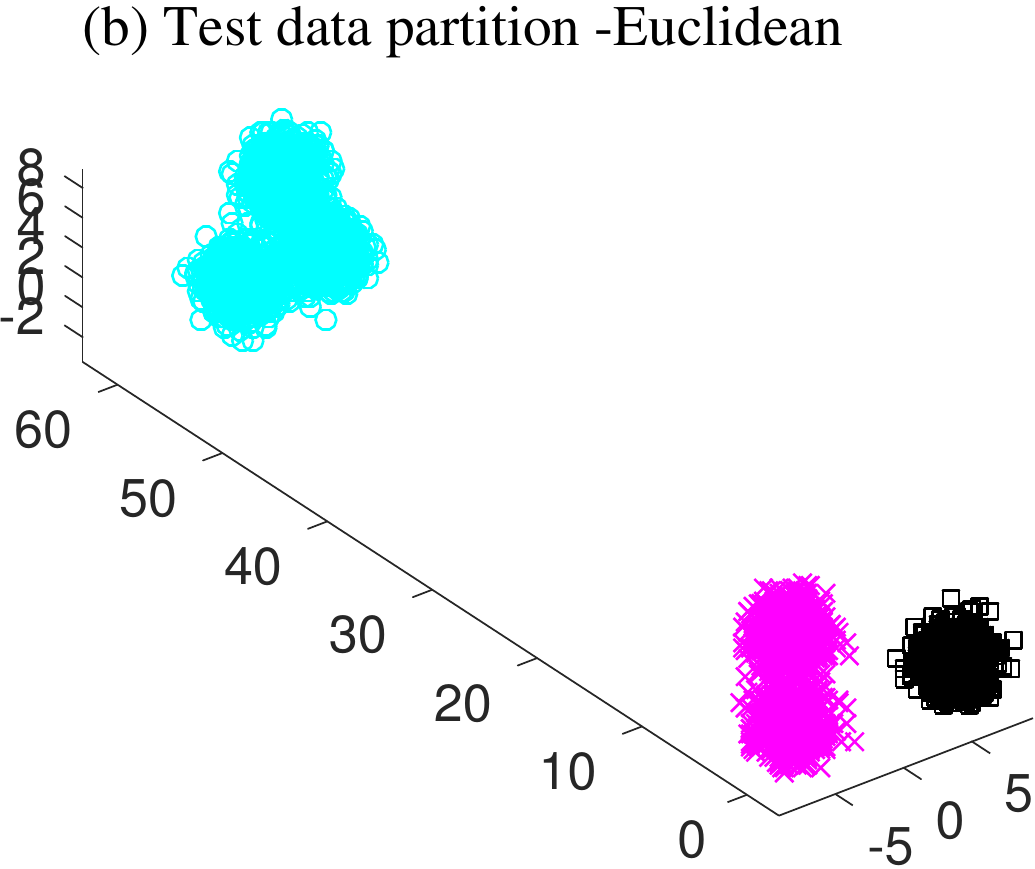}
\label{fig:disctSize}
}%
{%
\includegraphics[width=4.0cm]{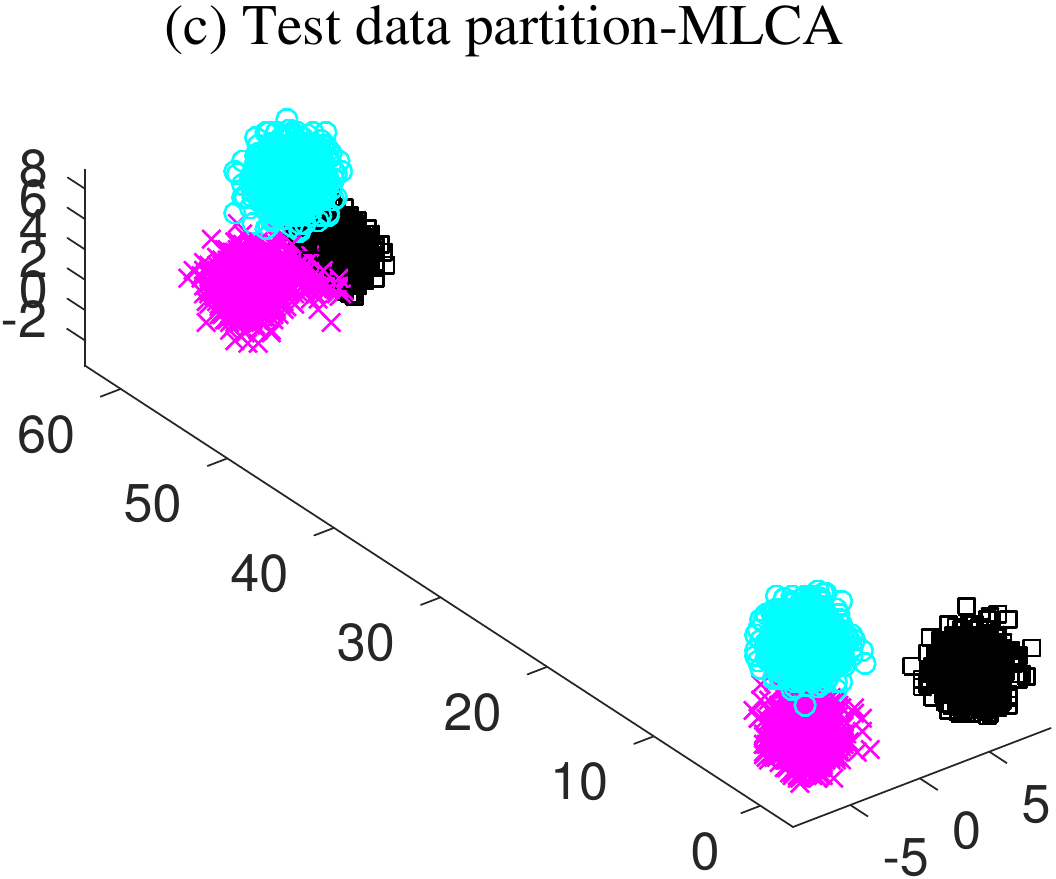}
\label{fig:disctSize}
}%
{%
\includegraphics[width=4.0cm]{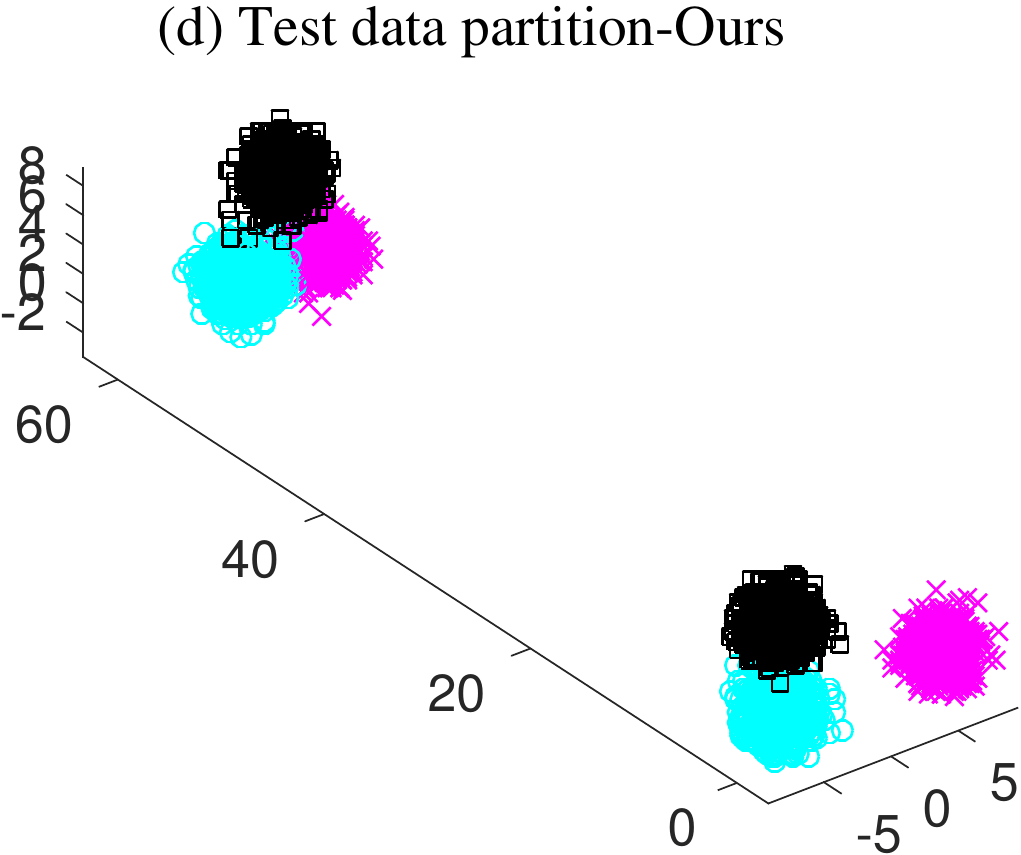}
\label{fig:disctSize}
}%

\caption{(a) Original training dataset, (b) Clustering obtained by $k$-means with Euclidean distance, (c) MLCA \cite{Law_2016_CVPR}, (d) SAE }
\label{sc_figure}
\end{figure*}


\noindent 
\textbf{Datasets}~~ Two datasets are used. A \textit{ synthetic dataset} is generated following \cite{Law_2016_CVPR}. Specifically, the training set is composed of 3-dimensional samples divided into 3 clusters, and each cluster has 1,000 samples.  Each of these clusters is composed of two subsclusters as shown in Fig.~\ref{sc_figure}(a). What makes the dataset difficult is that the subclusters of the same cluster are closer to the subsclusters from different categories than to each other when the distance is measured with Euclidean distance. Furthermore, some samples are corrupted by  noise which put them  in the subclusters of other categories in the feature space. We generate our test dataset with the similar properties (and the same number of examples N=3000) as the training set. To make clustering more challenging, the number of samples for each cluster is made different: 1000, 2000, and 4000 for three clusters respectively. This dataset is designed to evaluate how robust the method is against  the size of clusters and its ability to avoid being biased by the largest category.  More details on the dataset can be found in \cite{Law_2016_CVPR,lajugie2014large,xing2003distance}.
We also test our algorithm with a \textit{real dataset} --  Oxford Flowers-17 (848 images) \cite{nilsback2006visual}.  We follow exactly the same settings of \cite{Law_2016_CVPR}. Specifically, a ground truth foreground/background segmentation is provided for every image. To extract features, first, images are resized with a height of 100 pixels, and SIFT and color features (Lab, RGB, and intensity) are extracted from each 8$\times$8 patch centred at every pixel, resulting a 135D feature vector for each pixel. Each image has about $10^4$ patches, and the data matrix for the whole dataset has about $2.2 \times 10^6$ rows -- this is thus a large-scale problem.  The dataset has 5 random split with 200 images for training, 30 for validation, and the rest for testing. 


\noindent
\textbf{Evaluation metric}~~
We calculate the clustering quality with a loss defined as $\Delta =\|\hat{C}-C\|^2$ \cite{lajugie2014large,Law_2016_CVPR}, where $C$ and $\hat{C}$ are ground truth and predicted clustering matrix (obtained using $k$-means) respectively. 

\noindent
\textbf{Competitors~~} We compare our method with the state-of-the-art methods which all formulate the supervised clustering problem as a metric learning problem. These include  
Xiang \etal \cite{xing2003distance}, Lajugie \etal \cite{lajugie2014large}, KISSME \cite{kostinger2012large}, ITML \cite{davis2007information}, LMNN \cite{weinberger2009distance}, and  MLCA \cite{Law_2016_CVPR}.

\noindent
\textbf{Comparative evaluation~~} Table \ref{sc_same_size} and Table
\ref{sc_diff_size} show the synthetic data results with and without noise
respectively. It can be seen that in terms of clustering accuracy, our
method is much better than all compared methods. On computational
cost, our model is   more expensive than MLCA but much better than all
others. Figure \ref{sc_figure} visualises the clustering results. On the real image segmentation data, Table
\ref{flower} compares our SAE with 
other methods. Again, we can see that SAE achieves the best clustering accuracy. The training time for SAE is 93 seconds, while MLCA is 39
seconds. Note that the data size is $2.2\times 10^6$, so both are very efficient. 

\section{Conclusion}

We proposed a novel zero-shot learning model based on a semantic
autoencoder (SAE). The SAE model uses very simple and computationally
fast linear projection function and introduce an
additional reconstruction objective function for
learning a more generalisable projection function. We demonstrate
through extensive experiments that this new SAE model outperforms
existing ZSL models on six benchmarks. Moreover, the model is
further extended to address the supervised clustering problem and again
produces state-of-the-art performance.

\section*{Acknowledgement}
The authors were funded in part by the European Research Council under the FP7 Project
SUNNY (grant agreement no. 313243).

{\small
\bibliographystyle{ieee}
\bibliography{zsl_elyor}
}

\end{document}